\documentclass[akbc,twoside,11pt,lettersize]{article}
\usepackage{authblk}
\usepackage{akbc}
\usepackage{comment}
\usepackage{times}
\usepackage{latexsym}
\usepackage{amsmath}
\usepackage{graphicx}
\usepackage{afterpage}
\usepackage{url}
\usepackage{enumitem}
\usepackage{amssymb}
\usepackage{algorithm}
\usepackage{breqn}
\usepackage[nocomma]{optidef}
\usepackage{algpseudocode}
\usepackage{float}
\usepackage{mathtools}
\usepackage{url}
\usepackage{booktabs}

\newcommand{\propara}{\textsc{ProPara}}
\newcommand{\cooking}{\textsc{npn-Cooking}}

\newcommand{\modelname}{\textsc{Dynapro}}

\title{Procedural Reading Comprehension with \\ Attribute-Aware Context Flow}
\author[1]{Aida Amini}
\author[1]{Antoine Bosselut}
\author[2]{Bhavana Dalvi Mishra} 
\author[1,2]{Yejin Choi}
\author[1,2]{Hannaneh Hajishirzi}
\affil[1]{University of Washington}
\affil[2]{Allen Institute for AI}
\affil[ ]{\textit {\{amini91, antoineb , yejin, hannaneh\}@cs.washington.edu}}
\affil[ ]{\textit {bhavanad}@allenai.org}

\ShortHeadings{}{}
\finalcopy % Uncomment for camera-ready version, but NOT for submission.
\begin{document}

{\centering
\maketitle
}

%{\centering 
%sa
%}
% \author{\name Aida Amini \email minton@ptolemy.arc.nasa.gov \\
%       \name Andy Philips \email philips@ptolemy.arc.nasa.gov \\
%       \addr NASA Ames Research Center, Mail Stop: 244-7,\\
%       Moffett Field, CA  94035 USA
%       \AND
%       \name Mark D. Johnston \email johnston@stsci.edu \\
%       \addr Space Telescope Science Institute,
%       3700 San Martin Drive,\\
%       Baltimore, MD 21218 USA
%       \AND
%       \name Philip Laird \email laird@ptolemy.arc.nasa.gov \\
%       \addr NASA Ames Research Center,
%       AI Research Branch, Mail Stop: 269-2,\\
%       Moffett Field, CA  94035 USA}

% For research notes, remove the comment character in the line below.
% \researchnote

\maketitle

\begin{abstract}

  Procedural texts  often describe processes (e.g., photosynthesis and cooking) that happen over entities (e.g., light, food).
     In this paper, we introduce an algorithm for procedural reading comprehension by translating the text into a general formalism that represents processes as a sequence of transitions over entity attributes (e.g., location, temperature). Leveraging pre-trained language models, our model obtains entity-aware and attribute-aware representations of the text by joint prediction of entity attributes and their transitions. Our model dynamically obtains contextual encodings of the procedural text exploiting information that is encoded about previous and current states to predict the transition of a certain attribute which can be identified as a span of text  or  from a pre-defined set of classes. Moreover, our model achieves state of the art results on two procedural reading comprehension datasets, namely \propara{} and \cooking.
    
    %% continuous events (e.g., absorbing water, refrigerating ingredient) that results in changes of values of the attributes of the entities that are participating in the procedural text. 
    %the flow of context between time ste, our model predicts entity attributes from spans of text and predicts   Our model obtains a procedural text representation to  encoding of the  predicts attributes of entities that are \aida{(1) leveraging the consistency in state transitions by jointly training on the transition of states in addition to the values of them and (2) biasing the action and state predictions for an entity at any given time step based on a latent representation of the world state. }, 
    %,\aida{a collection of paragraphs describing scientific processes Our model outperform the state-of-the-art model by 2 points of f1 score. }
\end{abstract}

\section{Introduction}
\label{sec:intro}
% Show the contributions clearly
% point the contributions clearly and then based on that write the intro to address those

\begin{figure}[t]
\centering
\setlength{\belowcaptionskip}{-30pt}
\includegraphics[width=0.9\textwidth]{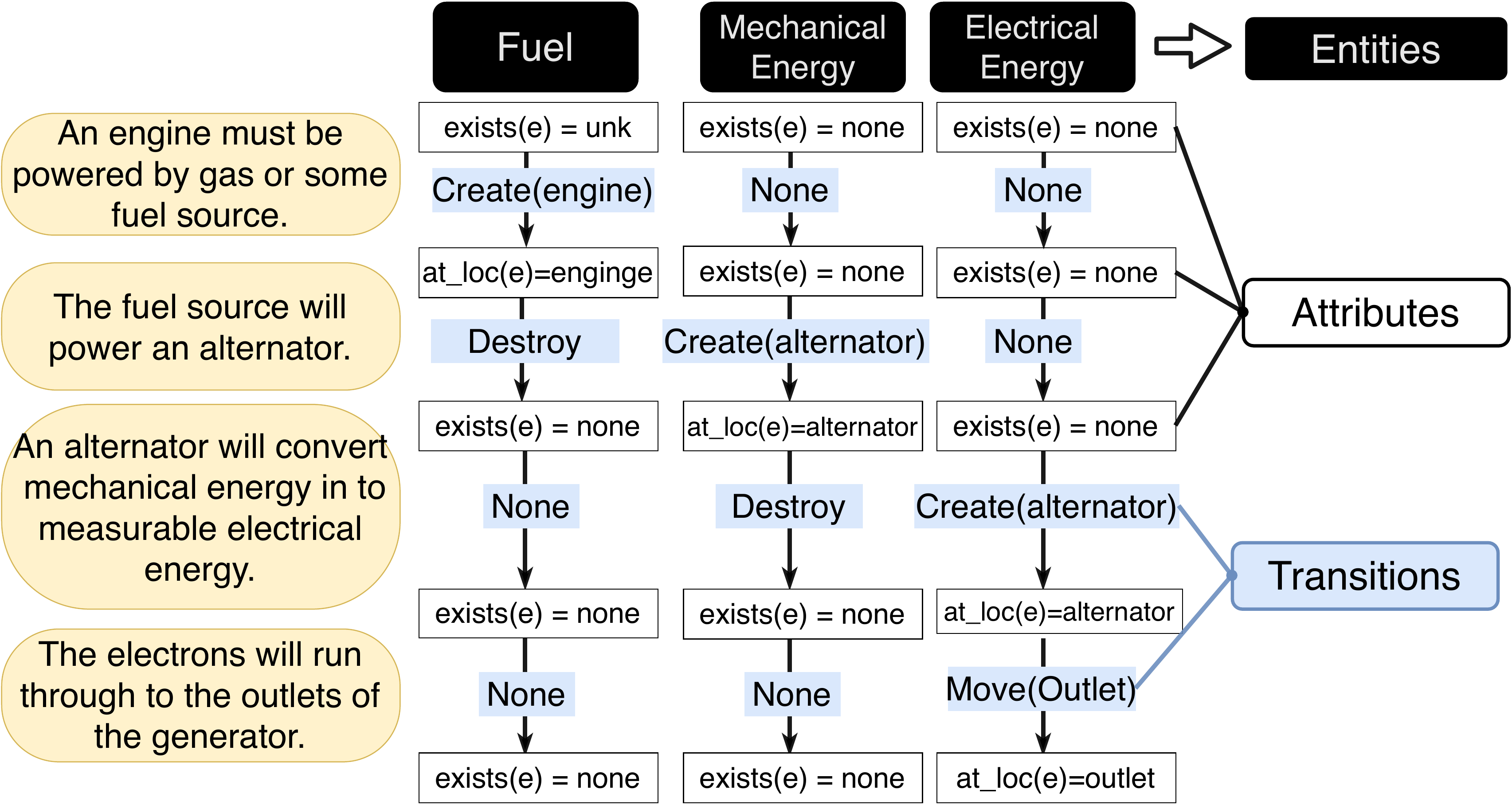}
\caption{Example of a procedural text and the predicted attributes and transitions for each entity.  Procedural reading comprehension is the task of answering questions about the underlying process. Sample questions in a \propara{} tasks are:  `What is the process input?',  `What is the process output?', `What is the location of the entity?'.} 
\label{fig:teaser}
\end{figure}

Procedural text describes how  entities (e.g., ${\tt fuel}$, ${\tt engine}$) or their  attributes (e.g., ${\tt locations}$)  change throughout a process (e.g., a { scientific process} or { cooking recipe}). Procedural reading comprehension is the task of answering questions about the underlying process in the text (Figure~\ref{fig:teaser}).  Understanding procedural text requires inferring  entity attributes and their dynamic transitions, which might only be implicitly mentioned in the text. For instance, in Figure~\ref{fig:teaser}, the  ${\tt creation}$ of the  
${\tt mechanical \ energy}$ in ${\tt alternator}$ can be inferred from second and third sentences. 

Full understanding of a procedural text requires capturing the full interplay between all components of a process, namely entities, their attributes and their dynamic transitions. 
Recent work in understanding procedural texts develop domain-specific models for tracking entities in scientific processes~\cite{Dalvi2018TrackingSC} or cooking recipes~\cite{Bosselut2018SimulatingAD}. More recently,   \citet{gupta-durrett-2019-entity-tracking} obtain general entity-aware representations of a procedural text leveraging pre-trained language models, and  predict entity transitions  from a set of pre-defined classes independent of entity attributes. Pre-defining the set of entity states limits the general applicability of the model since  entity attributes can be arbitrary spans of text. Moreover, {entity attributes} can be exploited for tracking entity state transitions.
% \eat{\aida{entity states -- I am suggesting replacing this with state transitions}, and identifying transition actions can help to better identity entity attributes and states. }
For example, in Figure \ref{fig:teaser}, the location of ${\tt fuel}$ can be effectively inferred from text as ${\tt engine}$ without the explicit mention of the ${\tt movement}$ transition in the first sentence.  Moreover,  the phrase ${\tt converted}$ in the third sentence gives rise to predicting two transition actions  of ${\tt destruction}$ of one type of ${\tt energy}$ and ${\tt creating}$ the other type.

In this work, we introduce a general formalism to represent procedural text and develop an end-to-end  neural  procedural reading comprehension model that jointly identifies entity attributes and  transitions leveraging dynamic contextual encoding of the procedural text. The formalism represents entities, their attributes, and their transitions across time. Our  model obtains attribute-aware representation of the procedural text leveraging a reading comprehension model that jointly identifies  entity attributes as a span of text or from a pre-defined set of classes. Our model predicts state transitions given the entity-aware and  attribute-aware encoding of the context up to a certain time step to consistently capture the dynamic flow of contextual encoding through an LSTM model. 

Our experiments show that our method achieves state of the art results across various tasks introduced on the \propara{} dataset to track entity attributes  and their transitions in scientific processes. Additionally, a simple variant of our model achieves state of the art results in the \cooking{} dataset. 

Our contributions are three-fold: (a) We present a general formalism to model procedural text, which can be adapted to different domains. (b) We  develop \modelname{}, an end to end neural model that  jointly and consistently predicts  entity attributes and their state transitions, leveraging pre-trained language models. (c) We show that our model can be adapted to several procedural reading comprehension tasks using the entity-aware and attribute-aware representations, achieving  state of art results on several diverse tasks.

\section{Related Work}\label{sec:related}
% \paragraph{Datasets} Reading comprehension is among the tasks that has drawn attention of researchers. There are large datasets on the task of question answering requiring reading a paragraph and follow the implicit and explicit fact. Among those we can name SQUAD(\cite{rajpurkar2016squad})and BABI(\cite{weston2015towards}) dataset.

Most previous work in reading comprehension~\cite{rajpurkar2016squad}  focus on identifying a span of text that answers a given question about a static paragraph. This paper focuses on procedural reading comprehension that inquires about how the states of entities change over time. Similar to us, there are several previous work that focus on understanding temporal text in multiple domains.  Cooking recipes~\cite{Bosselut2018SimulatingAD} describe instructions on how ingredients  consistently change.  bAbI~\cite{weston2015towards} is a collection of datasets focusing on understanding narratives and stories. Math word problems~\cite{kushman-etal-2014-learning, verbcat,amini2019mathqa,koncel2016mawps} describe how the state of entities change throughout some mathematical procedures. Narrative question answering~\cite{kovcisky2018narrativeqa,lin2019reasoning} inquires  to reason about the state of a story over time. The   \propara{} dataset~\cite{Dalvi2018TrackingSC} is a  collection of  procedural texts that describe how entities change throughout scientific processes over time, and inquires about several aspects of the process such as the   entity attributes or state transitions. Several models (e.g.,  EntNet\cite{Henaff2017TrackingTW}, QRN~\cite{Seo2017QueryReductionNF}, MemNet~\cite{weston2014memory}) have also been introduced to track entities in narratives.

%Tracking  entities and transitions is also explored with other tasks and datasets. The first one is EntNet, a network for tracking entities that was tried on children story book and bAbI and it was retried on \propara{} as well. The second one is about story generation by keeping track of the main entities(people) in the story.
% \citet{Henaff2017TrackingTW} used a memory network to maintain the entity states in  Children's Book Test. \citet{clark2018neural} show that modeling  entities and the changes that happens improves the quality of the story generations. 

The closest work to ours is the line of work focusing on the \propara\ and \cooking{} datasets.  \citet{Bosselut2018SimulatingAD} use an attention- based neural network to find  transitions in ingredients.  Pro-local and Pro-Global~\cite{Dalvi2018TrackingSC} first identify locations of entities using an entity recognition approach and use manual rules or global structure of the procedural text to  consistently track entities. \citet{Tandon2018ReasoningAA} leverage manually defined and knowledge-base driven commonsense constraints to avoid nonsensical predictions in Pro-Struct model (e.g., entity  trees don't moves to different locations).  KG-MRC~\cite{Das2018BuildingDK} maintain a knowledge graph of entities over time and identify entity states by predicting the location span with respect to each entity while utilizing a reading comprehension model. NCET (\citet{Gupta2019TrackingDA}) introduces a neural conditional random field model to maintain the consistency of state predictions. Most recently, ${ET_{BERT}}$\cite{gupta-durrett-2019-entity-tracking} uses transformers to construct entity-aware representation of each entity and predict the state transitions from a set of predefined classes. In this paper, we integrate all previous observation and develop a model that jointly identifies entities, attributes, and transitions over time. Unlike previous work that is designed to address either attributes or transitions, our model benefits from the clues that are implicitly and explicitly mentioned for both entity attributes and transitions. Leveraging both aspects of procedural reading comprehension lead us to a general and adaptive definition and model for such task that has achieved state of art in several tasks.   

\label{sec:background}
\section{Procedural Text Representation}
\label{Formalism}

Procedural text is a sequence of sentences describing how entities and their attributes change throughout a process. We introduce a general formalism to represent a procedural text:
\begin{equation}
p = (E, A, T),     
\end{equation} 
where $E$ is the list of entities participating in the process, $A$ is the list of entity attributes, and  $T$ is the list of transitions.

\paragraph{Entities} are the main elements participating in the process. For example, in the scientific processes described in  \propara{}  entities include elements such as ${\tt energy}$, ${\tt fuel}$, etc. In the cooking recipe domain, the entities are ingredients such as ${\tt milk}$, ${\tt flour}$, etc. The entities can be given based on the task such as in \propara{} and cooking domain or they can be inferred from the context (e.g., math word problems).  

\paragraph{Attributes} are  entity properties that can change over time. We model attributes as functions ${\tt Attribute(e)=val}$ that assign a value ${\tt val}$ to an attribute of the entity $e$. The entity state at each time is derived by combining all the attribute values of that entity. Attribute values can be either spans of text or can be derived from a predefined set of classes.  For example, in \propara{} an important attribute of an entity is its ${\tt location}$ which can be a span of text. Npn-Cooking dataset introduces several  attributes (such as ${\tt shape}$ and ${\tt cookedness}$) for each ingredient. Example attributes addressing the entities in \propara{} are modeled as follows:
\begin{align*}
\begin{split}
& \tt{exists}(e)=\{\tt{none},\tt{unknown}, \tt{span in text}\}  \\
& \tt{at\_loc}(e)=l \rightarrow \mbox{\tt Assigns the location $l$ to entity $e$}
\end{split}
\end{align*} 
% \begin{equation}
% \mbox{at\_loc(e, l)} \rightarrow \mbox{True if the location of entity e is l}
% \end{equation}

% \paragraph{States} indicate the assignment of values to all attributes of entities at each time step, and shows the truth value of all the predicates. For all attributes $a_i$ over the entity $x$  \hanna{define the following equation carefully, what are a, what is x, what are y, etc?}

% \begin{equation}
%     S_{t}(x) = [a_{1}(x, [y]_{|M|}) \land ... \land a_{N}(x, [y]_{|M|}) ]
% \end{equation}

\paragraph{Transitions} capture changes in the entity states. More specifically, transitions indicate how entity attributes change over time. We model each transition with an action name and a list of arguments that include the entity and some attribute values. For example, \propara\ consists of  four transition types :  $\tt{Create}(e, \tt{loc})$, $\tt{Destroy}(e)$, $\tt{None}(e)$ and $\tt{Move}(e, \tt{loc})$. 

% \hanna{frog?}
% Figure \ref{fig:formalism} portraits the state of three different entities in the process of \textit{Frog's cycle of life}. At every time step different actions can happen to entities and the states are calculated accordingly.

% \begin{figure}[t]
% \centering
% \includegraphics[width=1\textwidth]{formalism_engine2.pdf}
% \caption{Example of a procedural text. Each entity has a location that can change throughout the paragraph.}
% \label{fig:formalism}
% \end{figure}
\section{Model}

\begin{figure}[t]
\centering
  \includegraphics[width=0.7\textwidth]{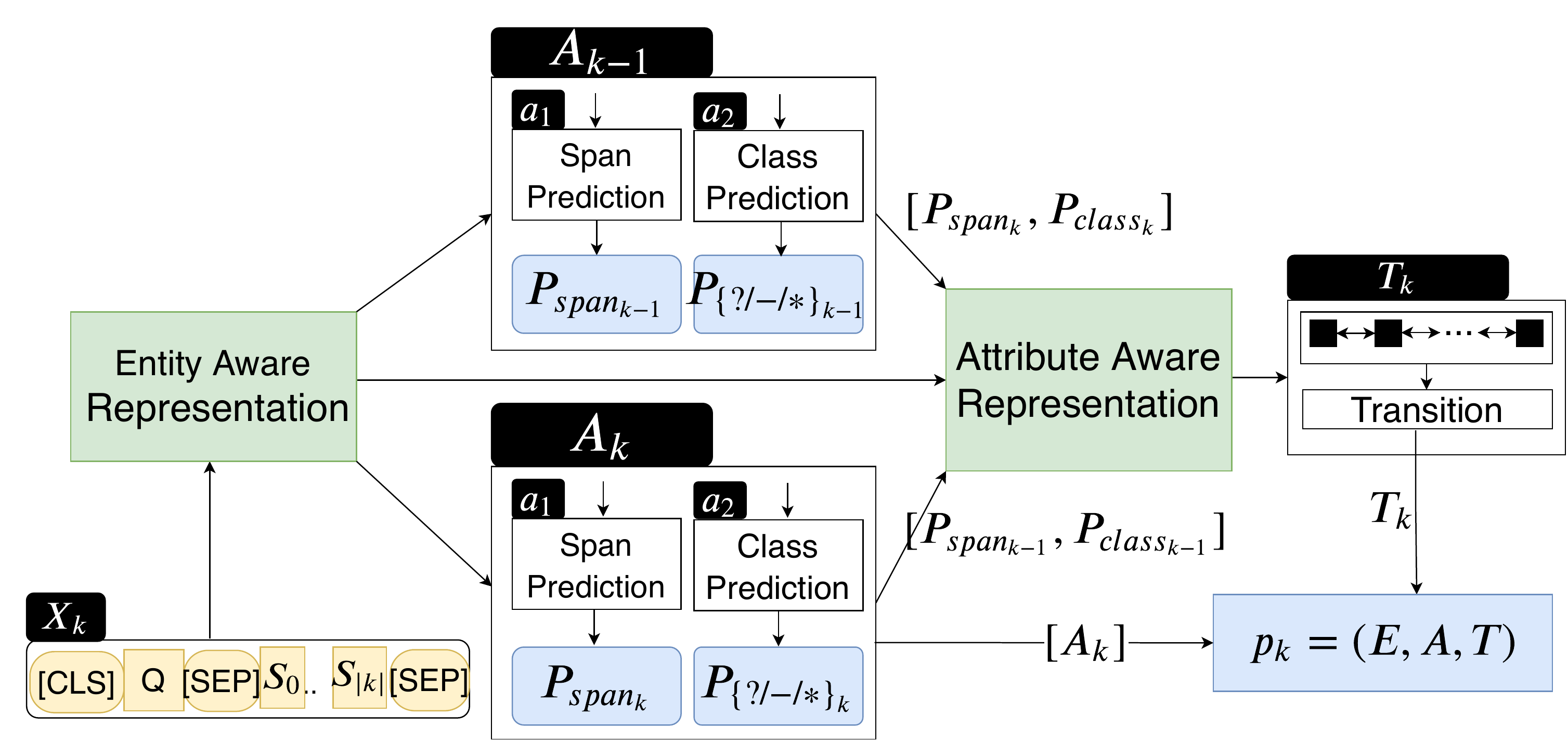}
 \setlength{\belowcaptionskip}{-30pt} 
 \caption{\modelname{} takes the procedural context $X_k$ as input and predicts attributes $A_{k-1}$, $A_k$  and  transitions $T_k$ at each time step $k$ $P_{\{?, -, *\}}$ indicates the probability of the location type among {\tt nowhere}, {\tt unkown}, and {\tt span of text} respectively.. The model uses the changes in attribute values from time steps $k-1$ to $k$ to predict transitions. } %The model predicts transitions using attribute-aware representation that captures from entity-aware representation  of the input   elements for Sequential entity Tracking.} %In this figure $K$ and $k-1$ shows the current and previous time steps respectively. $?$ shows the ${\tt unknown}$ location classification, $-$ shows ${\tt non-existent}$ class and $*$ shows {\tt somewhere within the span}}
    \label{fig:dynapro}
\end{figure}

% \subsection{Model Definition}
We introduce \modelname, an end-to-end neural architecture that jointly predicts entity attributes and their transitions.  Figure~\ref{fig:dynapro} depicts an overview of  our model.  \modelname{} first obtains the representation of the procedural text corresponding to an entity at each time step (Section~\ref{sec:entity-aware}). It then identifies   entity attributes for current and previous time steps (Section~\ref{sec:attribute}) and uses them to develop an attribute-aware representation of the procedural context (Section~\ref{sec:attribute-aware}). Finally, \modelname{}  uses entity-aware and attribute-aware representations to predict transitions that happen at that time step (Section~\ref{sec:transitions}). 
%%%%%%%%%%%%%%%%%%%%%%%%

\subsection{Entity-aware representation} \label{sec:entity-aware}
Given a procedural text $\langle S_0\ldots S_k \ldots S_T\rangle$  and an entity $e$, \modelname{} encodes procedural context $X_k$ at each time step $k$  and obtains the entity-aware representation vector $R_{k}(e)$. The procedural context is formed by concatenating entity name, query, and a fragment of the procedural text. The entity name and the query are included in the procedural context to capture the entity-aware representation of the context.  Since entity attributes are changing throughout the process, we form the context at each step $k$ by truncating the procedural text up to the $k^{th}$ sentence.  More formally, the procedural context is defined as: 
\begin{equation}
     X_k(e)=[cls]  Q_e [sep]  [C_i]  S_0 \ldots S_k   [sep], 
\end{equation}
where [$S_0 \ldots S_k$] is the fragment of the procedural text up to the $k^{th}$ sentence, $Q_e$ is the entity-aware query (e.g., ``Where is $e$?''), $[C_i]$ includes tokens that are reserved for attribute value classes (e.g., ${\tt nowhere}, {\tt unknown}$), and $[cls]$ and $[sep]$   are special tokens to capture sentence representations and separators. 

\modelname{} then uses a pre-trained language model to encode the procedural context $X_k(e)$ and returns the entity-aware representation $R_k(e)=BERT(X_k(e)).$ Hereinafter, for the ease of notation we will remove the argument $e$ from the equations.

%%%%%%%%%%%%%%%%%%%%%%%%%%%%%%%%%%

\subsection{Attribute Identification} \label{sec:attribute}
\modelname{} identifies  attribute values for each entity from the entity-aware representation $R_{k}(e)$ by jointly predicting attribute values from a pre-defined set of classes or extracts them as a text span. 

\paragraph{Class Prediction} Some attribute values can be identified from a set of pre-defined classes.  For instance ${\tt existence}$ attribute of an entity can be identified from $\{ {\tt nowhere}, {\tt unknown},{\tt span of text}\}$. Our model predicts the probability distribution $P_{class_k}$ over different classes of attribute values given the entity-aware representation $R_k$. 
\begin{equation}\label{eq:class}
 %   &R_{{seq}_k} = LSTM(h, R_{e_k})\\
    P_{class_k} = \mathrm{softmax}(f_{\theta_1}(g(R_k))), 
\end{equation}
where $R_k$ is the entity-aware representation, $g$ is a non-linear function, $f$ is a linear function and $\theta_{1}$ are  learnable parameters.

\paragraph{Span Prediction} Defining all attribute values apriori limits the general applicability of procedural text understanding. Some attribute values  are only mentioned within ${\tt span of text}$. For example, the  ${\tt location}$ of an entity may be mentioned in the text, but not as a set of pre-defined classes. For span prediction, we follow the standard procedure of phrase extraction in reading comprehension~\cite{seo2016bidirectional}  that predicts  two probability distributions over start and end tokens of the span. %\aida{Based on the task the sequential layer may be used for smoothing the probability distribution over start and end of spans -- should we remove this?} 
\begin{equation} \label{eq:span}
\begin{split}
 %   &R_{{seq}_k} = LSTM(h, R_{e_k})\\
    &P_{span_k} = [P_{start_k}, P_{end_k}] \\
    &P_{start_k} = \mathrm{softmax}(f_{\theta_2}(g(R_k)))\\
    &P_{end_k} = \mathrm{softmax}(f_{\theta_3}(g(R_k))),
\end{split}
\end{equation}
where $g$ is a non-linear function,  $f$ is a linear function and ${\theta_2}$ and ${\theta_3}$ are the learnable parameters used to find the probability distributions of start and end tokens of the span.

% \begin{equation}\label{eq:2}
% R_{out_{enc}}, h^{enc}_i = LSTM(h^{enc}_{i-1}, R_{i-1})
% \end{equation}

In order to capture the transitions of entity attributes, our model captures attributes for time steps $k-1$ and $k$ given a procedural context $X_k$. More specifically, we use equations~\ref{eq:class} and ~\ref{eq:span} to compute the probability distributions $P_{class_{k-1}}$,  $P_{span_{k-1}}$, $P_{class_{k}}$ and $P_{span_{k}}$ for both  time steps $k$ and $k-1$.

%%%%%%%%%%%%%%%%%%%%%%%%%%%%%%%%%%%%%%%%%%%%%%%%%

\subsection{Attribute-aware Representation}\label{sec:attribute-aware}
\modelname{} obtains attribute-aware representation $R_{a_k}$ of the context to encode entity attributes and their transitions at each time step $k$ using the predicted distributions $P_{span_k}$ and $P_{class_k}$ for each entity $e$. The intuition is to assign higher probabilities to the  tokens corresponding to the attribute value of the entity at time step $k$. %\aida{ How about this: We attend the probability distribution of each class and span of the attributes to the relating tokens to construct attention representation vector relating the entity and the attribute. } 

\begin{equation}
% \begin{split}
R_{a_k} =\sum_{class}(R_k. P_{class_k} \cdot m_{class})\cdot P_{span_k}(w),
% &R_{c_e} =\sum_{c_i \in classes}(R * P_{c_i} * m_{c_i}) * P_{c_e}(w) \\
% &R_{p_b} =\sum_{c_i \in classes}(R * P_{c_i} * m_{c_i}) * P_{p_b}(w) \\
% &R_{c_e} =\sum_{c_i \in classes}(R * P_{c_i} * m_{c_i}) * P_{c_e}(w) \end{split}
\end{equation} Where $class\in \{ {\tt nowhere}, {\tt unknown},{\tt span}\}$ are the predefined classification of attributes, $P_{class_k}$ and $P_{span_k}$  denote the probability distribution of attribute values over predefined classes and the span of text respectively, and are calculated using equations \ref{eq:class} and \ref{eq:span}.  $m_{class}$ is a vector that masks out the input tokens that do not correspond with a specific class. 
% where $class\in \{ {\tt nowhere}, {\tt unknown},{\tt span}\}$, $P_{class_k}$ and $P_{span_k}$ are calculated using equations \ref{eq:class} and \ref{eq:span}, and  show the probability distribution of attribute value over predefined classes and the span respectively, and $m_{class}$ is  a vector to mask out the input tokens that do not correspond to the specific class.

We model the  flow of the context by concatenating  attribute-aware representations for time step $k$ and $k-1$ as, 
\begin{equation}
    R_{a_{k-1:k}} = [R_{a_k}, R_{a_{k-1}}].
\end{equation}

%%%%%%%%%%%%%%%%%%%%%%%%%%%%%%%%%%%%%%%%%%%%%%%%%%
\subsection{Transition classification} \label{sec:transitions}
\modelname\ predicts attribute transitions from entity-aware and attribute-aware representations. In order to make smooth transition predictions and avoid redundant transitions we include a Bi-LSTM layer before the classification of the transition.
\begin{equation}\label{eq:transition}
\begin{split}
    &R_{{seq}_k}= \mathrm{LSTM}(h, [R_k,R{a_{k-1:k}}])\\
    &P_{transition_k} = \mathrm{softmax}(f_{\theta_4}(g([R_k,R_{{seq}_k}]))),
\end{split}
\end{equation}
where $h$ is the hidden vector of sequential layer, $\theta_4$ is the learnable parameter and $R_{{seq}_k}$ is the  output of the sequential layer.

\subsection{Inference and Training}
\paragraph{Training} Our model is trained end-to-end by optimizing the loss function below:
\begin{equation}
    loss_{total} = (loss_{span}, loss_{class})_{k-1} + (loss_{span}, loss_{class})_{k} + loss_{transition_{k}}
\end{equation}

Each loss function is defined as a cross entropy loss. $(loss_{span}, loss_{state})_{k}$ and $loss_{transition_k}$ are the losses of attribute prediction and the transition prediction modules at time step $k$, respectively.

% \bhavana{Can you clarify training instance creation? Maybe $A_{K}$, $A_{K-1}$ can be better explained as before and after states for step K. Right now someone might infer that A_k-1 is final output for previous step.}

\paragraph{Inference} 
At each time step $k$, the attributes $A_k$ and transitions $T_k$ are predicted given $P_{span_k}$, $P_{class_k}$, and $P_{transition_k}$. The final output  of the model consists of two sets of predictions, the attributes $A_{0\ldots k}$ and transitions $T_{0\ldots k}$ which are combined to track entities throughout a process given a task-specific objective (more in implementation details).

\label{sec:model}

\label{sec:ex_res_table}
\section{Experiments and Results}

\subsection{Datasets} 
We evaluate our model over the \propara{} dataset introduced by \cite{Dalvi2018TrackingSC} with the vocabulary size of 2.5k. This dataset contains over  400 manually-written paragraphs of scientific process descriptions. Each paragraph includes average of 4.17 entities and 6 sentences. The entities are extracted by experts and the transitions are annotated by crowd-workers. 

We additionally evaluate our model on the \cooking{} dataset. This corpus contains 65k cooking recipes. Each recipe consists of ingredient tracked during the process. Training samples are heuristically annotated by string matching and dev/test samples are annotated by crowd-workers. We randomly sample from the training recipes that have contained ingredients which changed in ${\tt location}$ attribute. 
\subsection{Tasks and metrics}\label{sec:metirc}
 We evaluate \modelname\ on three tasks in \propara{} and one task in \cooking{}.

\paragraph{Document-level predictions}  This task is introduced by \citet{Dalvi2018TrackingSC} that evaluates four different objectives per entity and process: Whether the entity is the (1) input or (2) output  of the process. (3) The moves and (4) the conversions of the entity in the process. The final metric reported for this evaluation is the average precision, recall and F1 score of all four questions. 

\paragraph{Sentence-level predictions}
The task is introduced by \citet{Dalvi2018TrackingSC} that considers questions about the procedural text: \textbf{Cat-1} asks if the specific entity is Created/Destroyed/Moved, \textbf{Cat-2} asks the time step at which the entity has been Created/Destroyed/Moved, and \textbf{Cat-3} asks about the location that entity is Created/Moved/Destroyed. The evaluation metric calculates the score of all transitions in each category and reports the micro and macro average of the scores among three categories.

\paragraph{Action dependencies} The task is recently introduced by \citet{Mishra2019EverythingHF} to check whether the actions predicted by a model have some role to play in overall dynamics of the procedural paragraph. The final metric reported for this task is the  precision, recall, and F1 scores of the dependency links averaged over all paragraphs.  
%\hanna{is this last sentence correct?}

\paragraph{Location prediction in Recipes}
The task is to identify the location of different entities in the cooking domain. In this domain, the list of attributes are fixed. We evaluate by measuring the change in location~\cite{Bosselut2018SimulatingAD} and compute F1 and accuracy in attribute prediction.  

%We evaluate \modelname{} on the location prediction task of the \cooking{} datasets. In order to fairly compare with \cite{Bosselut2018SimulatingAD}, we measure the change in location and compare it with their f1 metric in state prediction task and compared state accuracy with accuracy of our predicted attributes.

%%%%%%%%%%%%%%%%%%%%%%%%%%%
\subsection{Implementation Details}
We use the official implementation of $BERT_{base}$ huggingface library~\cite{Wolf2019HuggingFacesTS}. We choose cross entropy loss function. The learning rate for training is $3 e^{-5}$ and the training batch size is $8$. The hidden size of the sequential layer is set to $1000$ and $200$ for class prediction and transition prediction respectively.

We  use the predicted $A_{k-1}$ to initialize the attribute of timestep 0 and at any other timestep we use at $A_k$ predictions for finding the value of attribute at timestep $k$. In the sentence level evaluation task introduced in \cite{Dalvi2018TrackingSC}, the consistency is not required. Inference phase for this task only uses the attribute predictions.
For the document-level predictions,  we construct the final predictions by favoring the transition predictions.  In case of inconsistency where there is no valid attribute prediction to support the transition we refer to the attribute value to deterministically infer the transition.

%To combine the final attribute values on \propara{} we favor the class prediction over the span prediction. If the class prediction indicates that the location is {\tt within span}, we refer to the selected span.

%The output of the model consists of two sets of predictions, the attributes $A_{0\ldots k}$ and transitions $T_{0\ldots k}$ which are combined to track entities throughout a process given a task-specific objective (more in implementation details). given a specific task These two can be combined together based on the task to answer .
%In inference time, the intitial attributes $A_0$ is predicted based on the attributes predicted for time step $0$. For the rest of the attribute values we use transition predictions.

To adapt the results of \modelname{} to identify action dependencies, we postprocess the results using similar heuristics described in the original task.
To adapt \modelname{} to the \cooking{} dataset, we use a $243$-way classification to predict attributes because the attributes are known apriori.  %\, we use the fix set of attribute classifications. In this domain, the defined transition are not varied from attributes. Therefore our adaptation model for this task uses a $243$-way classification layer for attribute  class prediction. The input is constructed in the sentence level manner. 

\subsection{Results and Analyses} 
Table \ref{table:question_results} and Table~\ref{table:cooking_results} compare \modelname\ with previous work (detailed in Section~\ref{sec:related}) in  \propara\  and \cooking\  tasks. As shown in the tables, \modelname\   outperforms the state of the art models in most of the evaluations.

\vspace{.1cm} \noindent{\bf Document-level task} We observe the most significant gain (3 absolute points in F1) in the document-level tasks, indicating the ability of the model in global understanding of the procedural text by joint predictions of entity attributes and transitions throughout time. Overall, \modelname\ predicts transitions with higher confidence, and hence results in high precision in most document-level tasks. 

\vspace{.1cm} \noindent{\bf Sentence-level task} 
\modelname\ outperforms the state-of-the-art models on Ma-Avg and Mi-Avg metrics  when comparing the full predictions and gives comparable results to previous work on change and time step predictions. Note that $ET_{BERT}$ \cite{gupta-durrett-2019-entity-tracking} only predicts actions (Create, Destroy, Move) but fails to predict location attributes as spans \modelname\  obtains a good performance on Cat-1 and Cat-2 prediction while learning to predict questions with more complex structure.  We   speculate that our lower numbers in {\bf Cat-1} and {\bf Cat-2} are due to \modelname's highly confident decisions that lead to high precision, but lower prediction rate, noting that {\bf Cat-1} and {\bf Cat-2} evaluate accuracy. %the lower recall compared\hanna{Why our model is worse in cat 1 and cat2? why higher confidence results in worse performance in cat1?} \aida{ I think it is because lower recall. The model predicts high precision, but the prediction rate is lower, the cat 1 is accuracy metric }

%\aida{\modelname{} tends to predict transition with higher confidence threshold (avoiding hallucinations) that results in high precision in \textbf{Document Level Task} but this has impacted the score of \textbf{Cat-1} predictions in \textbf{Sentence Level Task.}}

\vspace{.1cm}\noindent {\textbf{Action Dependency}}
 \modelname{}   outperforms all previous work   with F1 score of 43.7.  Note that XPAD \cite{Mishra2019EverythingHF} explicitly favors predicting state changes that result in dependencies across steps. In contrast, \modelname{} is only optimized to track entities. 

\vspace{.1cm}\noindent {\textbf{Location prediction in Recipes}} Finally, a simple variant of \modelname{} achieves best performance in the \cooking{} dataset, showcasing the importance of procedural text encoding over time. 

% Table \ref{table:question_results} shows that 

\begin{table}[t]
\small
\centering
\begin{tabular}{l|ccccr|ccc|ccc}
\toprule
%& &&  Sentence Level&&& & Document Level \\
& \multicolumn{5}{c}{Sentence-Level} & \multicolumn{3}{c}{Document Level}& \multicolumn{3}{c}{Action Dependency} \\
Model       &  Cat-1 & Cat-2 & Cat-3 & Ma-Avg & Mi-Avg &  P & R &  F$_1$ & P & R & F$_1$ \\
\toprule
ProLocal    &  62.7 & 30.5 & 10.4 &  34.5 &  34.0 & \textbf{77.4} & 22.9 & 35.3 &24.7&18.0& 20.8\\
EntNet      &  51.6 & 18.8 & 7.8  &  26.1 &  26.0  & 50.2 & 33.5 & 40.2 &32.8&38.6& 35.5 \\
QRN         &  52.4 & 15.5 & 10.9 &  26.3 &  26.5 & 55.5 & 31.3 & 40.0  &32.6&30.3& 31.4\\
ProGlobal   &  63.0 & 36.4 & 35.9 &  45.1 &  45.4 & 46.7 & 52.4 & 49.4 &43.4&37.0& 39.9  \\
KG-MRC      &  62.9 & 40.0 & 38.2 &  47.0 &  46.6 & 64.5 & 50.7  & 56.8 & 46.5& 39.5	&42.7  \\
NCET        &  70.6 & 44.6 & 41.3 &  52.2 &  52.3 & 64.2 & 53.9 & 58.6 &-&-& -  \\
NCET + ELMo &  \textbf{73.7} & 47.1 & 41.0 &  53.9 &  54.0  & 67.1 & \textbf{58.5} & 62.5 &  50.4 &	28.6 &	36.5\\
$ET_{BERT}$&  73.6 & \textbf{52.6} & - &  - &  -  &-&-& -&-&-& - \\
XPAD &-&-& -&-&-& 70.5&45.3&55.2&  62.0 &	32.9	&	43.0\\
\midrule
\modelname   &     72.4  & 49.3 & \textbf{44.5}  &   \textbf{55.4}     &     \textbf{55.5}   & 75.2 & 58.0 & \textbf{65.5} & \textbf{64.9}&	\textbf{32.9}	&\textbf{43.7} \\
\bottomrule
\end{tabular}
\caption{Results comparing \modelname~to previous state of the art methods on {sentence-level }, {document-level } and {Action Dependency} tasks of \propara{}  (test set).}
\label{table:question_results}
\vspace{-2mm}
\end{table}

\begin{table}[t]
\centering
\begin{tabular}{l|cc} 
Model  &  F1  & Accuracy \\ \hline
NPN-cooking &35.1 & 51.3 \\
KG-MRC &- & 51.6\\
 \hline
\modelname & \textbf{36.3} & \textbf{62.9}\\

\end{tabular}
\caption{F1 and accuracy score on the location prediction task in \cooking{}.}
\label{table:cooking_results}
\end{table}

\subsection{Ablation Studies and Analyses}
In order to better understand the impact of \modelname's components, we evaluate different variants of \modelname{} in the document-level task of the \propara{} dataset.  
\begin{itemize}[noitemsep] 
    \item {\bf No attribute-aware representation} The model only considers entity-aware representations in Equation~\ref{eq:transition} for transition predictions. 
    \item {\bf No transition classification} The model does not include  transition classification. 
    \item {\bf No sequential modeling} The model removes the sequential smoothing of the predicted transitions by removing LSTM from Equation~\ref{eq:transition}.
    \item {\bf No class prediction} The model only uses span predictions. 
    \item {\bf CLS instead of attribute-aware representation} The model uses the [cls] encoding of $R_k$ instead of the attribute-aware representation $R_{a_k}$.
    \item {\bf Full procedural input} that uses the full text of the procedure instead of the truncated text $X_k$ at time step $k$.
\end{itemize}

Table~\ref{table:ablation_results} shows that removing each component from \modelname\ hurts the performance, indicating that joint prediction of attribute spans, classes, and transitions are all important in procedural reading comprehension. Moreover, the table shows the effect of attribute-aware representations that incorporate the flow of context by predicting attributes of two consecutive time steps. Finally, the table shows the effect of procedural context modeling by truncating sentences up to a certain time step rather than considering the full document at each time step. Note that document-level evaluation in \propara{} requires  spans of texts being identified, therefore removing span prediction from \modelname\ cannot be ablated.

\begin{table}[t]
\centering
\begin{tabular}{lc} 
Ablation  &  F1  \\ \hline
Full model (\modelname) &  \textbf{71.9}\\
\hline
%- sequential span prediction &  \textbf{71.9}\\
% - sequential state classification & 70.10 \\
No attribute aware representation & 69.5 \\
No transition prediction & 66.3\\
No sequential modeling in transition module  & 68.8\\ 
No sequential modeling in attribute classification  & 68.2\\ 
No class prediction & 53.8 \\
CLS instead of attribute-aware representation & 70.9\\
Full procedural input  &  61.0\\ \hline
\end{tabular}
\caption{Ablation study of different components in \modelname\ by comparing F1 score on \propara{} \textbf{Document Level task}   (dev set).}
\label{table:ablation_results}
\end{table}

%\label{sec:experiments}
%\subsection{\cooking{} Dataset Evaluation} 
% \textbf{Action dependencies:} We  evaluate \modelname{} on the new task of explaining the actions by predicting their dependencies\cite{Mishra2019EverythingHF}. This task checks whether the actions predicted by a model have some role to play in overall dynamics of the procedural paragraph.  Table \ref{table:F1-dependency-task} shows that \modelname{} results in SOTA performance with F1=43.69. Note that XPAD explicitly favors predicting state changes that result in dependencies across steps. On the other hand, \modelname{} is optimized to predict accurate state changes, and its results are later post-processed to predict action dependencies using the heuristic method described in \cite{Mishra2019EverythingHF}.

% \section{Experimental Results}
% \label{sec:exp_res}
% \subsection{Ablations}
\

\label{sec:ex_res}

\subsection{Error Analysis}
\paragraph{Qualitative Analyses} Table \ref{table:q_exxample} shows the three types of common mistakes in the final predictions. In the first example \modelname{} successfully tracks the ${\tt blood}$ entity while circulating in the ${\tt body}$, yet there is a mismatch of what portion of the text it chooses as the span. In the second example, the model correctly predicts the location of ${\tt carbon dioxide}$ as ${\tt blood}$, but there is not enough external knowledge provided for the model to predict that this entity gets ${\tt destroyed}$. In the third example, the model mistakenly predicts  the ${\tt air and petrol}$ as a container for the ${\tt energy}$, and since the changes are explicitly happening to the container they are not propagate to the ${\tt entity}$. 

\begin{table*}[t]
\setlength{\belowcaptionskip}{20pt}
\centering
\begin{tabular}{lllll} 
\# & Sentence & Gold  & Prediction\\ \hline
% Vapor is condensed out in the form of water droplets. & water droplets   & air  & air \\
1.1 & Blood enters the \underline{right side of your \textbf{heart}}. & heart  & right side of your heart\\
1.2 & Blood travels to the \underline{\textbf{lungs}}. & lungs & lungs \\
1.3 & Carbon dioxide is removed from the blood &  lungs & lungs \\
1.4 & Blood returns to \underline{left side of your \textbf{heart}}  & heart  & left side of your heart \\ \hline
2.1 & \underline{\textbf{Blood}} travels to the lungs & blood  & blood \\
2.2 & Carbon dioxide is removed from the blood. & - & ? \\ \hline
3.1 & Fuel converts to energy when \underline{air and petrol} mix.  & -  & air and petrol \\
3.2 & The car \textbf{engine} burns the mix of \underline{air and petrol}. & engine  & air and petrol \\
3.3 & Hot gas from the burning pushes the \textbf{pistons}. & piston & air and petrol \\
3.4 & The resulting energy powers the \underline{\textbf{crankshaft}}. & crankshaft  & crankshaft  \\\hline
\end{tabular}
\caption{Examples of correct and incorrect predictions of \modelname{}.  Entities in the first, second, and third examples are ${\tt blood}$, ${\tt carbon dioxide}$, ${\tt energy}$, respectively.}
\label{table:q_exxample}
\end{table*}

\vspace{-.3cm}
\paragraph{Inconsistent Transitions} We categorize possible inconsistencies in transition predictions into three categories. (The percentage numbers shows how many times that inconsistency was observed in the inference step.):
\begin{itemize}[noitemsep] 
    \item \textbf{Creation(2.0\%):} When the supporting attribute is predicted to be ${\tt non}$-${\tt existence}$ or the previous attribute shows that the entity already ${\tt exists}$.
    \item \textbf{Move(1.5\%):} When the predicted attribute is not changed from previous prediction or it refers to a ${\tt non}$-${\tt existence}$ case. 
    \item \textbf{Destroy(1.0\%):} When the predicted attribute for the last timestep is ${\tt non}$-${\tt existence}$.
\end{itemize}

\label{sec:analysis}
\section{Conclusion}
We introduce an end-to-end model that benefits from both entity-aware representations and attribute-aware representations to jointly predict attributes values and their transitions related to an entity. We present a general formalism to model procedural texts and introduce  a model to translate procedural text into that formalism. We show that entity-aware and temporal-aware construction of the input helps to achieve better entity-aware and attribute-aware representations of the procedural context. Finally, we show how our model can achieve inferences about state transitions by tracking transition in attribute values.  Our model achieves the state of the art results on various tasks over the \propara{} dataset and the \cooking{} dataset. Future work involves extending our method to automatically identifying entities and their attribute types and adapting to other domains.

\label{sec:conclusion}

\newpage

\bibliography{sample}
\bibliographystyle{plainnat}

\end{document}